\documentclass[10pt, a4paper]{article}
\usepackage{lrec}
\usepackage{multibib}
\newcites{languageresource}{Language Resources}
\usepackage{graphicx}
\usepackage{tabularx}
\usepackage{soul}

\usepackage{epstopdf}
\usepackage[utf8]{inputenc}

\usepackage{hyperref}
\usepackage{xstring}

\usepackage[vietnamese,english]{babel}
\usepackage{color}

\newcommand{\comment}[1]{#1}

\title{BKTreebank: Building a Vietnamese Dependency Treebank}

\name{Kiem-Hieu Nguyen}

\address{School of information and communication technology, \\
		Hanoi university of science and technology, \\
			1 Dai Co Viet, Bach Khoa, Hai Ba Trung, Hanoi, Vietnam \\
         hieunk@soict.hust.edu.vn}

\abstract{
Dependency treebank is an important resource in any language. In this paper, we present our work on building BKTreebank, a dependency treebank for Vietnamese. Important points on designing POS tagset, dependency relations, and annotation guidelines are discussed. We describe experiments on POS tagging and dependency parsing on the treebank. Experimental results show that the treebank is a useful resource for Vietnamese language processing. \\ 
\newline 
}

\begin{document}

\maketitleabstract

\section{Introduction}

Dependency treebank is important for data-driven dependency parsing. However, building a dependency treebank is complicated and expensive.

Dependency treebanks have been available in English and many languages. VnDT \cite{Nguyen2014} is a Vietnamese dependency treebank which was automatically converted from tree bracketing in VietTreebank (VTB) \cite{nguyen-EtAl:2009:LAW-III,Nguyen:2015:VTC:2812480.2812510}. 

In this work, we present the building of a dependency treebank for Vietnamese\footnote{For information on using BKTreebank, please visit \\ \url{http://is.hust.edu.vn/~hieunk/bktreebank/}}. Our treebank was manually annotated by annotators. Its annotation guidelines substantially differ from VTB. Our contributions are two-fold:
\begin{itemize}
	\item	A manual dependency treebank for Vietnamese.
	\item	Experiments on POS tagging and dependency parsing based on the treebank.
\end{itemize}

The paper is organized as follows: Section~\ref{section:related_work} briefly introduces related work on building treebanks for Vietnamese and dependency treebanks for other languages. Section~\ref{section:guidelines} highlights important points of annotation guidelines. Section~\ref{section:annotation} describes in brief the annotation process. Section~\ref{section:evaluations} is dedicated to evaluations and discussions on automatic POS tagging and dependency parsing results. The paper is concluded in Section~\ref{section:conclusions}

\section{Related Work}
\label{section:related_work}

\subsection{Treebanks for Vietnamese}

VTB was the pioneer treebank for Vietnamese. It has been developed from 2006-2010. It contains manual annotations on about 40K sentences for word segmentation, 10K sentences for POS tagging, and 10K sentences for bracketing.

VnDT contains dependency annotations which were automatically converted from bracketing annotations in VTB. State-of-the-art performance on VnDT is 80.7\% and 73.5\% on UAS and LAS, respectively \cite{QuocNguyen:Dras:Johnson:2016:ALTA2016}.

Recently, a new treebank for Vietnamese has been developed \cite{NGUYEN16.95,Nguyen2017}. It consists of 40K sentences annotated with word segmentation, POS tagging, and bracketing. While generally agreeing on word segmentation and bracketing, they propose a POS tagset and POS tagging guidelines which focus more on word-class transformation, particularly between verbs and other word-classes. This issue is important as Vietnamese is an analytic language. Unfortunately, their treebank has not been publicly available for research community yet. 

\subsection{Dependency treebank for other languages}

One of the most notable dependency treebanks for English was developed by Stanford NLP group \cite{deMarneffe:2008:STD:1608858.1608859}. The Stanford treebank is automatically converted from PeenTreebank phrase structures \cite{marneffe:440:2006:lrec2006}. Similar approaches were used to build dependency treebanks in other languages such as French, Korean, and Croatian \cite{CANDITO10.392,choi-park-choi:2012:SP-SEM-MRL,BEROVIC12.719}. Other treebanks are built manually for languages such as Norwegian \cite{SOLBERG14.303}.

The Universal Dependencies is inherited from Penn POS tagset and Stanford typed dependency representation, and has been expanded to many languages \cite{DEMARNEFFE14.1062,NIVRE16.348}.

\section{Annotation Guidelines}
\label{section:guidelines}

\subsection{POS tagging guidelines}

\selectlanguage{vietnamese}
Our POS tagset relies on Penn tagset \cite{santorini93penntagset} with the following adaptation to Vietnamese (see Table~\ref{table:pos} for the full tagset):
\begin{itemize}
\item As Vietnamese is an analytic language, we omit tags related to plurality, tense, and superlative in Penn tagset. 
\item \textit{CL} is used for noun classifiers. In Vietnamese, a countable noun could be accompanied by a classifier when we want to indicate quantity or simply to emphasize. For example, `tấm' is a classifier' in ``Anh ta giành được hai tấm huy chương vàng'' (He won two gold medals); `chiếc' is a classifier in ``Chiếc xe này khá đắt'' (This car is quite expensive). In \cite{NGUYEN16.95}, the authors also dedicate two tags \textit{Nc} and \textit{Ncs} for noun classifiers. Similar phenomena could be found in other languages such as Korean \cite{Kim2006}.
\item \textit{PFN} is used for prefix nominalizers. Many nominal expressions in Vietnamese are formed by a leading nominalizer and a verb or an adjective (see Table~\ref{table:nominalization} for examples). In \cite{NGUYEN16.95}, there are also POS tags mentioning word-class transformation including VA (Verb-Adjective), VN (Verb-Noun), and NA (Noun-Adjective) but it is not clear from the paper how the tags are designed.
\item \textit{NML} is used for phrasal nominalizers. In Vietnamese, a special word such as `việc' is used as a clausal adverbial marker for a clausal component. For instance, in ``Việc xử lý chất thải công nghiệp cần được làm ngay'' (The processing of industry garbage needs to be done immediately), `việc' is the marker for the clausal subject.
\item \textit{VA} is used for adjectival verb. In Vietnamese, when the predicate is an adjective, there is no copula verb \textit{to be}. It is hence tagged as an adjectival verb. In the sentence ``Tình hình tương đối khả quan'' (The situation is\footnote{Note that there is no \textit{to be} in the sentence in Vietnamese due to \textit{zero copula}.} quite positive), `khả quan' is predicate and is tagged as VA.
\item \textit{AV} stand for verbal adjective. When a verb modifies a noun, it is tagged as an verbal adjective (e.g. biển/NN quảng\_cáo/AV (advertising board)). 
\item \textit{TO} is used to tagged `để', which has similar meaning as ``in order to'' in English.
\end{itemize}


\selectlanguage{english}

\begin{table}[!ht]
\begin{center}
\selectlanguage{vietnamese}
\begin{tabular}{|l|l|l|}
      \hline
		\textbf{Prefix nominalizer} & \textbf{Word} & \textbf{Expression} \\ 
      \hline\hline
		niềm & vui & niềm vui (happiness) \\
		sự & hi sinh & sự hi sinh (sacrifice) \\
		niềm & tin & niềm tin (belief) \\
      \hline
\end{tabular}
\selectlanguage{english}
\caption{Examples of prefix nominalizer in Vietnamese.}
\label{table:nominalization}
 \end{center}
\end{table}

\begin{table}[!h]
\begin{center}
\selectlanguage{vietnamese}
\begin{tabular}{|l|l|}
      \hline
      \textbf{POS tag} & \textbf{Description} \\
      \hline\hline
		CD & Cardinal number \\
      	DT & Determiner \\
      	MD & Modal \\
      	NN & Noun \\	
      	NNP & Proper noun \\
      	NML* & Phrasal nominalizer \\	
      	PFN* & Prefix nominalizer \\	
      	PRP & Personal pronoun \\	
      	RB & Adverb \\
      	VB & Verb \\	
      	VA* & Adjectival verb \\	
      	IN & Preposition \\
      	JJ & Adjective \\	
      	AV* & Verbal adjective \\	
      	PUNCT & Punctuation \\
      	CC & Coordinating conjunction \\	
      	WDT & Wh-determiner \\
      	WP & Wh-pronoun\\	
      	WRB & Wh-adverb \\
      	CL* & Noun classifier\\	
      	TO & `để' (in order to) \\
      	UH & Interjection \\	
		FW & Foreign word \\
      \hline
\end{tabular}
\selectlanguage{english}
\caption{Our POS tagset (* Tag specific for Vietnamese).}
\label{table:pos}
 \end{center}
\end{table}

\subsection{Dependency parsing guidelines}
\selectlanguage{vietnamese}
Our dependency relations relies on Stanford dependencies \cite{deMarneffe:2008:STD:1608858.1608859} (Table~\ref{table:dependency}). We add two relations for norminalization:
\begin{itemize}
\item \textit{case:pfn} is used for nominalizing modifier between a headword as a nominalizer and a verb or an adjective (see examples in Table~\ref{table:nominalization}).
\item \textit{mark:relcl} is used for phrasal adverbial modifier between a headword as the predicate of the clause and a marker such as `việc'. 
\end{itemize}
Guidelines for other relations are similar to Stanford dependencies with some modifications. For instance,  
\begin{itemize}
\item \textit{aux} is also used for relationship between a verb and a tense auxiliary (e.g. thực hiện/VB - aux - đang/MD in ``đang thực hiện'' (be executing)).
\item \textit{det} is also used for relationship between a noun and its plural marker. Here, we tag a plural marker as a determiner (e.g. trường hợp/NN - det - những/DT in ``những trường hợp'' (cases)).
\end{itemize}
\selectlanguage{english}
\begin{table}[!h]
\begin{center}
\selectlanguage{vietnamese}
\begin{tabular}{|l|l|}
      \hline
      \textbf{Relation} & \textbf{Description} \\
      \hline\hline
		nsubj & Nominal subject \\
		nsubjpass & Passive nominal subject \\
		dobj & Direct object \\
		iobj & Indirect object \\
		csubj & Clausal subject \\
		csubjpass & Passive clausal subject \\
		ccomp & Clausal component \\
		xcomp & Open clausal component \\
		advcl & Adverbial clause modifier \\
		advmod & Adverbial modifier \\
		aux & Auxiliary \\
		cop & Copula \\
		mark & Marker \\
		mark:relcl* & Phrasal nominalizer \\
		nmod & Nominal modifier \\
		appos & Appositional modifier \\
		nummod & Numeric modifier \\
		acl & Adjectival clause \\
		amod & Adjectival modifier \\
		det & Determiner \\
		case:pfn* & Prefix nominalizer \\
		case & Case marking \\
		conj & Conjunct \\
		cc & Coordinating conjunction \\
		punct & Punctuation \\
		dep & Unspecified dependency \\
      \hline
\end{tabular}
\selectlanguage{english}
\caption{Our dependency relations (* Relation specific for Vietnamese)}
\label{table:dependency}
 \end{center}
\end{table}

\section{Annotation Process}
\label{section:annotation}

\begin{figure*}[!ht]
\begin{center}
\includegraphics[width=\textwidth]{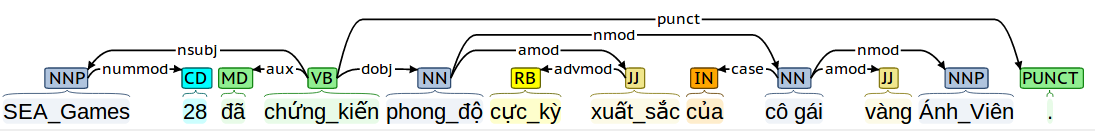} 
\caption{An annotation example in BRAT (SEA Games 28 witnesses an excellent performance from Golden Girl Anh Vien).}
\label{figure:brat}
\end{center}
\end{figure*}

The raw corpus was collected from Dantri\footnote{\url{http://dantri.vn}}, a general-domain online news agency.

Texts were first segmented by UETSegmenter \cite{7800279}. Sentences longer than 50 words were removed. Three annotators produced manual POS tagging and dependency parsing using the annotation tool BRAT \cite{stenetorp-EtAl:2012:DemoEACL2012}.


We decided to annotate POS tagging and dependency in parallel because the two tasks are complimentary to each other. After being explained the annotation guidelines, the annotators were first asked to separately annotate the same small sample dataset. After finishing the sample dataset, they discussed differences and agreed on final decisions.

After being trained, each annotator were asked to annotate separate documents. They discussed with each other when dealing with confusing cases. Every week, the annotators together reviewed and discussed a random annotated document. In the final round, a forth annotator reviewed all annotations and discussed with the annotators in the previous round when necessary to make final decisions.

After removing invalid parsed sentences, our treebank contains \comment{6909} manually annotated sentences on POS tagging and dependency parsing with the average speed of 7 min/sentence. 

Figure~\ref{figure:brat} illustrates an annotation example using BRAT. Segmented texts are put into BRAT. Syllables of the same word are connected by `\_'. POS tags are labeled for each words. 

\section{Annotation Evaluations}
\label{section:evaluations}
\subsection{Inter annotator agreement}

After finishing annotation, the three annotators were asked again to separately annotate the same small dataset to measure Inter-Annotator-Agreement (IAA). Averaged \textit{kappa} is \comment{94.5, 85.2, and 80.4} for POS tagging, unlabeled dependency parsing, and labeled dependency parsing, respectively. Note that IAA was measured for separate annotations of the three annotators without revising of the forth one. Such agreement shows good coherence between different annotators.

\subsection{Initial results on POS tagging and dependency parsing}

The treebank was divided into a training set of 5639 sentences and a test set of 1270 sentences for learning and testing POS tagging and dependency parsing. 

We built a vanilla POS tagging model using CRFSuite\footnote{\url{http://www.chokkan.org/software/crfsuite/}} implementation of first-order Conditional Random Fields with default hyper-parameters. We used a straightforward feature set as described in Table~\ref{table:POS_features}. Our lexicon was built by merging the lexicon of VietTreebank \cite{10.2307/30208384} with frequent tags in our corpus considering important differences in tagging guidelines. Only (word, tag) pairs that were tagged more than three times in the corpus were considered and were reviewed before adding to the lexicon. 

\begin{table}[!h]
\begin{center}
\begin{tabular}{|l|r|}
      \hline
      \textbf{Feature set} \\
      \hline\hline
		w[-2], w[-1], w[0], w[1], w[2] \\
		candidate tags \\
		is\_head\_capitalized \\
		is\_all\_capitalized \\
		is\_numeric \\
      \hline
\end{tabular}
\caption{Feature set for learning POS tagger with CRF}
\label{table:POS_features}
 \end{center}
\end{table}

\begin{table}[!h]
\begin{center}
\begin{tabular}{|l|r|r|r|}
      \hline
      \textbf{Tag} & \textbf{P} & \textbf{R} & \textbf{F} \\
      \hline\hline
		NN & 92.4 & 93.6 & 93.0 \\
		IN & 89.0 & 95.0 & 91.9 \\
		MD & 97.6 & 98.3 & 98.0 \\
		VB & 89.6 & 91.1 & 90.3 \\
		VA & 58.2 & 41.6 & 48.6 \\
		CD & 89.1 & 97.7 & 93.2 \\
		RB & 84.2 & 87.0 & 85.5 \\
		CL & 85.3 & 71.1 & 77.6 \\
		AV & 59.4 & 42.3 & 49.4 \\
		PUNCT & 99.9 & 100.0 & 99.9 \\
		JJ & 85.9 & 66.9 & 75.2 \\
		NNP & 91.9 & 94.5 & 93.2 \\
		DT & 97.1 & 94.6 & 95.8 \\
		PFN & 73.9 & 86.7 & 79.8 \\
		CC & 92.5 & 96.3 & 94.4 \\
		PRP & 90.7 & 88.6 & 89.7 \\
      \hline
		\multicolumn{4}{|l|}{\textbf{Overall accuracy: 90.7}} \\
		\hline
\end{tabular}
\caption{\comment{POS performance by tag}}
\label{table:POS_performance_per_tag}
 \end{center}
\end{table}

We used the transition-based MaltParser \cite{nivre_hall_nilsson_chanev_eryigit_kubler_marinov_marsi_2007} with default algorithm and feature set\footnote{\url{http://www.maltparser.org/userguide.html}} to built a vanilla dependency parser.

\begin{table}[!h]
\begin{center}
\begin{tabular}{|l|r|r|}
      \hline
      \textbf{Relation} & \textbf{UAS} & \textbf{LAS} \\
      \hline\hline
		ROOT & 80.4 & 80.4 \\
		acl & 63.4 & 63.4 \\
		advcl & 64.7 & 39.5 \\
		advmod & 86.8 & 86.2 \\
		amod & 89.9 & 89.4 \\
		aux & 98.4 & 97.4 \\
		auxpass & 98.8 & 92.8 \\
		case & 97.5 & 97.5 \\
		case:pfn & 100.0 & 100.0 \\
		cc & 84.9 & 84.9 \\
		ccomp & 77.9 & 46.8 \\
		cl & 100.0 & 100.0 \\
		conj & 59.9 & 48.9 \\
		cop & 95.4 & 94.2 \\
		csubj & 75.0 & 63.9 \\
		dep & 72.2 & 72.2 \\
		det & 97.1 & 97.1 \\
		dobj & 92.4 & 89.2 \\
		mark & 93.1 & 93.1 \\
		mark:relcl & 100.0 & 100.0 \\
		neg & 89.6 & 85.6 \\
		nmod & 78.2 & 74.3 \\
		nsubj & 86.2 & 79.6 \\
		nsubjpass & 93.5 & 75.8 \\
		nummod & 91.6 & 89.5 \\
		punct & 73.9 & 73.7 \\
		xcomp & 79.9 & 70.9 \\
		\hline
		\textbf{Overall} & \textbf{84.4} & \textbf{81.4} \\
      \hline
\end{tabular}
\caption{\comment{Dependency parsing performance by relation}}
\label{table:dependency_parsing}
 \end{center}
\end{table}

\subsection{Discussions}

As shown in Table~\ref{table:POS_performance_per_tag}, performance of POS tagging on nouns is similar to averaged performance. Verbs are more difficult to tag as they are ambiguous, not only with nouns and adjectives, but also with verbal adjective (modifiers). Automatic tagging of verbal adjective modifiers is very challenging as such modifiers are not infectional, and in some cases it requires knowledge at syntactic level. They are usually mistakenly tagged as a predicate verb. Verbal adjectives are also difficult because of zero-copula phenomenon.

Dependency parsing performance is promising as shown in Table~\ref{table:dependency_parsing}. Parsing at phrase-level is accurate except for nominal modifiers perhaps due to confusing usage of directional and temporal adverbial nouns and prepositions in Vietnamese. On the other hand, parsing at clause-level is poor. There are plenty rooms for improvement on such long-distance dependencies.

\section{Conclusion}
\label{section:conclusions}

In this paper, we present the building of a dependency treebank for Vietnamese. Our work is based on previous works on treebanks for Vietnamese and dependency treebanks for other languages. Although current size of the corpus is limited, initial experimental results on POS tagging and dependency parsing is promising.

In the future, we are going to expand BKTreebank with a bootstrapping approach using automatic tagger and parser learned from the dataset. We are going to investigate several approaches to POS tagging and dependency parsing for Vietnamese, including the joint learning approach. 

\section{Acknowledgements}

This project has been partially funded by VCCorp via collaboration with Data science laboratory, School of information and communication technology, Hanoi university of science and technology. We would like to thank Vu Xuan Luong for enthusiastic discussions on VietTreebank.

\section{Bibliographical References}
\label{main:ref}

\bibliographystyle{lrec}
\bibliography{references}


\end{document}